%% file: main.tex
\documentclass{article} % DO NOT CHANGE THIS]
\usepackage[hyphens]{url}  % DO NOT CHANGE THIS
\usepackage{graphicx} % DO NOT CHANGE THIS

\usepackage[utf8]{inputenc}
\usepackage{amssymb,amsthm,amsmath,mathrsfs,mathtools,thmtools,thm-restate}
\usepackage{comment}
\usepackage{appendix}
\allowdisplaybreaks

\usepackage{xspace}
\usepackage{bigstrut}

\newcommand{\magone}{\textsc{Mag-L1}\xspace}
\newcommand{\magtwo}{\textsc{Mag-L2}\xspace}

\usepackage{tabu}
\newtheorem*{theorem*}{Theorem}

\theoremstyle{definition}

\theoremstyle{remark}

\title{A Comparative Study of Neural Network Compression}

\author{Hossein Baktash\thanks{Contributed to the paper during a summer internship in INRIA Sophia Antipolis, France, Funded by I3S Labratoire, CNRS.}\\
Sharif University of Technology, Iran
\and
{Emanuele Natale}\\
CNRS, UCA, I3S, INRIA, France
\and 
{Laurent Viennot}\\
INRIA, IRIF, France 
}

\begin{document}
    \maketitle
\begin{abstract}
    \input{trunk/abstract.tex}
\end{abstract}

\input{trunk/intro.tex}

\input{trunk/related.tex}
\input{trunk/compression.tex}
\input{trunk/experiments.tex}
\input{trunk/discussion.tex}

\newpage
\bibliography{biblio}
\bibliographystyle{alpha}

\end{document}

%% file: trunk/abstract.tex
There has recently been an increasing desire to evaluate neural networks locally on computationally-limited devices 
in order to exploit their recent effectiveness for several applications; 
such effectiveness has nevertheless come together with a considerable increase in the size of modern neural networks, 
which constitute a major downside in several of the aforementioned computationally-limited settings. 
There has thus been a demand of \emph{compression techniques} for neural networks. 
Several proposal in this direction have been made, which famously include 
hashing-based methods and
pruning-based ones. 
However, the evaluation of the efficacy of these techniques 
has so far been heterogeneous, with no clear evidence
in favor of any of them over the others. 

The goal of this work is to address this latter issue 
by providing a comparative study. 
While most previous studies test the capability of a technique in 
reducing the number of parameters of state-of-the-art networks, 
we follow \cite{chen_compressing_2015} in evaluating their performance 
on basic architectures on the MNIST dataset and variants of it, 
which allows for a clearer analysis of some aspects of their behavior. 
To the best of our knowledge, we are the first to directly compare
famous approaches such as HashedNet, Optimal Brain Damage (OBD), and magnitude-based pruning
with L1 and L2 regularization among them and against equivalent-size feed-forward
neural networks with simple (fully-connected) and structural
(convolutional) neural networks. 
Rather surprisingly, our experiments show that (iterative) pruning-based methods 
are substantially better than the HashedNet architecture, whose compression doesn't 
appear advantageous to a carefully chosen convolutional network. 
We also show that, when the compression
level is high, the famous OBD pruning heuristics deteriorates to the point of being
less efficient than simple magnitude-based techniques. 

%% file: trunk/intro.tex
\section{Introduction}

Because of their increasing use in applications over mobile devices and embedded systems, 
there has been recent surge of interest in developing methods to reduce the size of neural networks \cite{hu_hashing_2018}.
The Moving Picture Experts Group, for example, recently opened a part of the MPEG-7 Standard in order to standardize 
aspects of neural networks compression for multimedia content description and analysis \cite{noauthor_compression_2019}. 
In the primary setting which is considered in this context, the client on which the networks has to be stored and
evaluated has weak storage and computational capabilities, while the server on which the network is trained can be 
assumed to be a high-performance cluster. 

The neural-network compression techniques explored so far can be summarily grouped in the following main families of strategies: 
matrix decomposition methods, such as low-rank decompositions \cite{liu_sparse_2015};
weight-sharing techniques, such as hashing-based and quantization schemes \cite{hu_hashing_2018};
pruning techniques, such as Optimal Brain Damage (OBD) \cite{le_cun_optimal_1989} and other iterative pruning-based methods. 

The motivation for this work stems from what appears to be a major gap in the literature encompassing
these techniques, as for a satisfying and methodical assessment of their relative efficacy. 
In fact, the series of works exploring these methods has been quite heterogeneous with respect to
the way they have been compared to each other. 
We thus aim for a principled comparison which provides substantial evidence in favor of some of them 
over the others. 
The main principle at the basis of our study is to strive for the most \emph{minimalist} setting: 
we consider simple architectures with a single hidden layer 
and perform our experiments on the classical MNIST dataset and variants of it \cite{larochelle_empirical_2007}, 
in opposition to most previous work which focused on reducing the number of parameters of state-of-the-art 
architectures.
Such choices allow for a clearer analysis of some aspects of the behavior of the technique.
A similar approach has been adopted in \cite{chen_compressing_2015} 
(with some limitations that we discuss in the Related Work section). 

We now briefly discuss the compression methods we consider. 
Further details are provided in the Related Work and technical section
of the paper.
We first remark that, while low-rank methods have been shown to achieve good speed-ups for large
convolutional kernels, they usually perform poorly for small kernels \cite{hu_hashing_2018}, 
and appear inferior to hashing-based methods \cite{chen_compressing_2015}.
We thus focus on the two other classes of techniques.
Hashing-based schemes, in particular, have attracted much attention. 
The central idea behind them is to employ hash functions to randomly group edges in 
\emph{buckets}, and then constrain edges belonging to the same bucket to share the same weight.
In this work, we focus on the main representative of this family of methods, 
namely the HashedNet architecture \cite{chen_compressing_2015} . 
The third family of techniques, (iterative) pruning methods, date back to the seminal work by LeCun et al. 
in 1989 on the Optimal Brain Damage (OBD) technique \cite{le_cun_optimal_1989}, 
and have recently regained attention \cite{to_prune_or_not_to,nervanasystemsdistiller}. 
For brevity's sake, we omit the adjective \emph{iterative} in the rest of the paper, 
since we don't consider non-iterative pruning methods. 
In pruning methods, the central idea is to sparsify the network during training by choosing some edges to be removed.
While in OBD and older methods the choice of edges to be removed is made according to heuristics 
that aims at minimizing the loss of the network accuracy, recent work has focused on magnitude-based
methods, in which the edges to be removed are simply those with the smallest absolute value.
Rather surprisingly, recent work do not validate their results against the aforementioned older methods. 

\subsubsection{Our Contribution}

To the best of our knowledge, this is the first work that provide an explicit comparison of compression approaches 
such as HashedNet, OBD and magnitude-based pruning with L1 and L2 regularization, which we also compare
against convolutional networks with number of parameters equivalent to that of the the final compressed network
(see the Discussion section where we argue that convolutional networks can be regarded as a structured weigh-sharing method). 
Rather surprisingly, our experiments show that pruning methods 
are substantially better than the HashedNet architecture, whose compression doesn't 
appear advantageous to a carefully chosen convolutional network.
Even more interestingly, magnitude-based pruning appear to perform at least as good as OBD, 
outperforming it consistently when the compression ratio\footnote{The compression ratio is defined 
as $r:=\frac{m'}{m} $ where $m$ is the number of parameters of the uncompressed network 
while $m$ is the number of parameter of the compressed one.} lead to very sparse topologies. 
It has been already noted in the literature that the assumptions on which the OBD heuristic 
is based on fail to hold in practice \cite{hassibi_optimal_1993}: our results show that 
this is true to the extent that magnitude-based edge-removal turns out to be a better heuristic
for minimizing the accuracy loss. 
The reader will find a discussion on the above results and further consequences of our experiments in the Discussion section. 
Overall, while surely not exhaustive, our work assess the relative efficacy of 
neural-network compression methods in a basic setting, and it provides fundamental insights on 
their shortcomings and strengths. 

\input{figs/old_mnist_convolution_vs_hashed.tex}
\input{figs/mnist_basic_100nhu.tex}
\input{figs/mnist_rotation_100nhu.tex}
\input{figs/mnist_background_random_100nhu.tex}
\input{figs/mnist_background_images_100nhu.tex}
\input{tables/MagL2_100nhu.tex}

\input{tables/MagL1_30nhu.tex}

\input{tables/MagL1_100nhu.tex}

\input{tables/MagL2_30nhu.tex}

\input{tables/OBD_100nhu.tex}

\input{tables/OBD_30nhu.tex}

\input{figs/hist_230.tex}

%% file: figs/old_mnist_convolution_vs_hashed.tex
\begin{figure}[t]
    \includegraphics[width=1\textwidth]{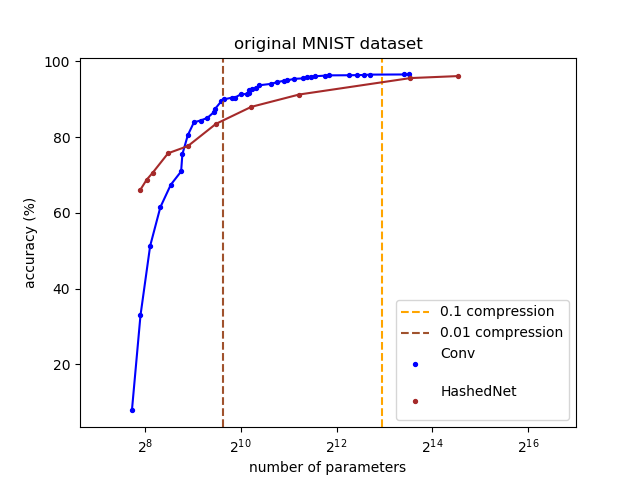}
    \caption{Accuracies of the HashedNet and convolutional network architectures, 
    based on the number of parameters, on the MNIST dataset. 
    The results for HashedNet are obtained by compressing a network with $100$ hidden units 
    in a single hidden layer from $0\%$ compression up to $99\%$ compression. 
    The convolutional networks consistently outperform HashedNet.}
    \label{fig:old_mnist_convolution_vs_hashed}
\end{figure}{}

%% file: figs/mnist_basic_100nhu.tex
\begin{figure}[ht!]
    \includegraphics[width=1\textwidth]{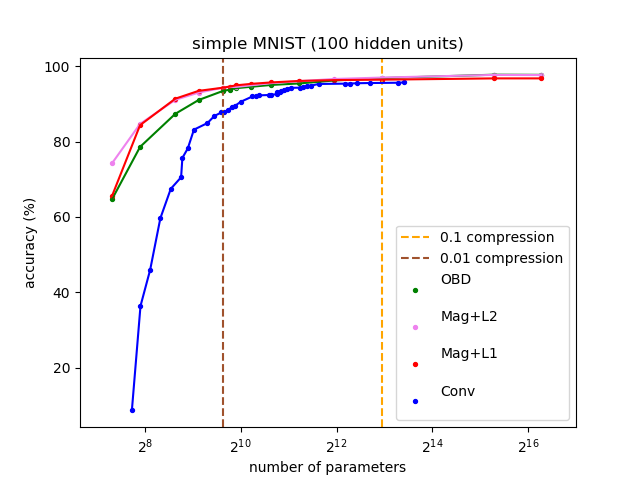}
    \caption{Accuracies of \emph{pruning} methods and simple convolutional networks, 
    with almost the same number of parameters on the MNIST dataset. 
    The results for the pruning methods are obtained by compressing 
    fully connected networks with 100 hidden units, 
    starting from $0\%$ compression up to $99\%$ compression. }
    \label{fig:mnist_basic_100nhu}
\end{figure}{}

%% file: figs/mnist_rotation_100nhu.tex
\begin{figure}[t]
    \includegraphics[width=1\textwidth]{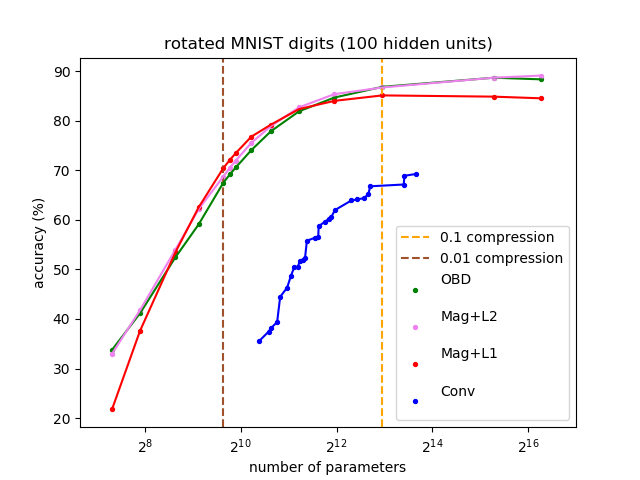}
    \caption{Accuracies of \emph{pruning} methods and simple convolutional networks 
    with almost the same number of parameters on the \emph{rotated} MNIST dataset. 
    The results are obtained by compressing fully connected networks with 100 hidden units, 
    starting from $0\%$ compression up to $99\%$ compression.}
    \label{fig:mnist_rotation_100nhu}
\end{figure}{}

%% file: figs/mnist_background_random_100nhu.tex
\begin{figure}[t]
    \includegraphics[width=1\textwidth]{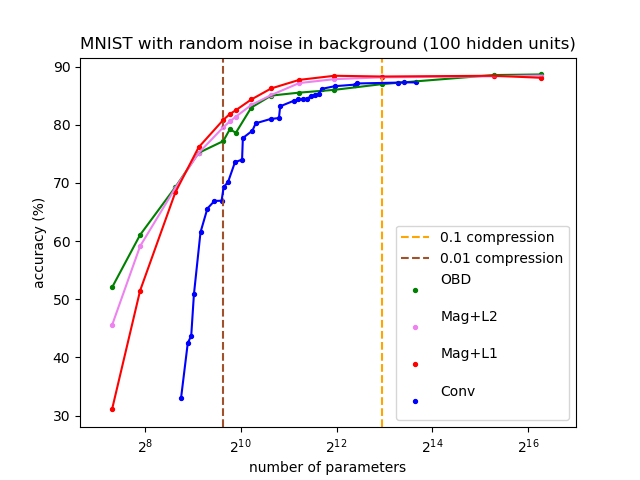}
    \caption{Accuracies of \emph{pruning} methods and simple convolutional network 
    with almost the same number of parameters on the MNIST dataset with random background.
    The results are obtained by compressing fully connected networks with 100 hidden units, 
    starting from $0\%$ compression up to $99\%$ compression. }
    \label{fig:mnist_background_random_100nhu}
\end{figure}{}

%% file: figs/mnist_background_images_100nhu.tex
\begin{figure}[t]
    \includegraphics[width=1\textwidth]{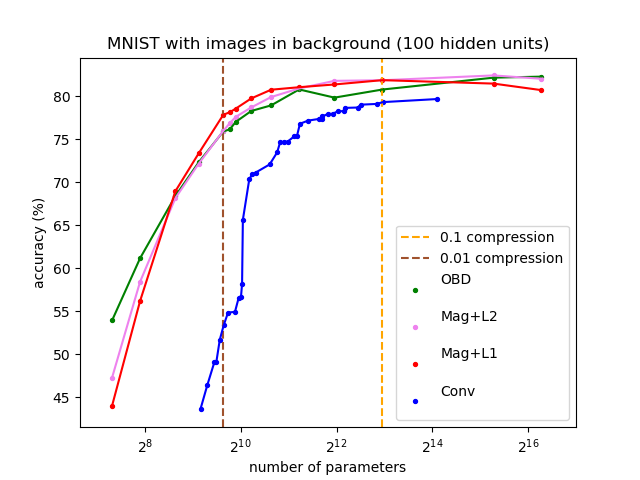}
    \caption{Accuracies of \emph{pruning} methods and simple convolutional network 
    with almost the same number of parameters on the MNIST dataset with images in background. 
    The results are obtained by compressing fully connected networks with 100 hidden units, 
    starting from $0\%$ compression up to $99\%$ compression.}
    \label{fig:mnist_background_images_100nhu}
\end{figure}{}

%% file: tables/MagL2_100nhu.tex
\begin{table}[t]
    \centering
    \begin{tabu} [ 0.4\textwidth] { | X[c] | X[c] | X[c] | X[c] | X[c] |}  
     \hline
     \multicolumn{5}{ | c | }{MagL2 100 hidden units one hidden layer}\\
     \hline
     edges removed & mnist & mnist background images & mnist background random & mnist rotation\\
     \hline
     $0$  $\%$   & $97.65_{\,0.11}$ & $82.04_{\,0.13}$ & $88.44_{\,0.23}$ & $89.16_{\,0.56}$\\ \hline
     $50$ $\%$   & $97.69_{\,0.05}$ & $82.43_{\,0.12}$ & $88.33_{\,0.38}$ & $88.72_{\,0.45}$ \\ \hline
     $90$ $\%$   & $96.99_{\,0.18}$ & $81.87_{\,0.54}$ & $88.15_{\,0.64}$ & $86.75_{\,0.37}$\\ \hline
     $95$ $\%$   & $96.62_{\,0.23}$ & $81.80_{\,0.41}$ & $87.84_{\,0.47}$ & $85.41_{\,0.58}$ \\ \hline
     $97$ $\%$   & $96.02_{\,0.25}$ & $80.92_{\,0.52}$ & $87.16_{\,0.41}$ & $82.75_{\,0.38}$\\ \hline
     $98$ $\%$   & $95.52_{\,0.13}$ & $79.92_{\,0.55}$ & $85.10_{\,0.18}$ & $79.00_{\,0.48}$ \\ \hline
     $98.5$ $\%$ & $94.94_{\,0.19}$ & $78.71_{\,0.66}$ & $83.40_{\,0.23}$ & $75.55_{\,0.21}$\\ \hline
     $98.8$ $\%$ & $94.50_{\,0.18}$ & $77.62_{\,0.36}$ & $81.38_{\,0.32}$ & $72.00_{\,0.50}$ \\ \hline
     $98.9$ $\%$ & $94.54_{\,0.16}$ & $76.89_{\,0.44}$ & $80.56_{\,0.39}$ & $70.51_{\,0.41}$\\ \hline
     $99$ $\%$   & $94.31_{\,0.17}$ & $75.98_{\,0.45}$ & $79.57_{\,0.34}$ & $68.76_{\,1.09}$ \\ \hline
     $99.3$ $\%$ & $93.00_{\,0.22}$ & $72.16_{\,1.04}$ & $75.14_{\,0.24}$ & $62.17_{\,1.55}$\\ \hline
     $99.5$ $\%$ & $91.09_{\,0.20}$ & $68.17_{\,0.41}$ & $69.15_{\,0.66}$ & $53.97_{\,0.39}$ \\ \hline
     $99.7$ $\%$ & $84.69_{\,1.01}$ & $58.45_{\,3.05}$ & $59.11_{\,2.19}$ & $41.80_{\,2.65}$\\ \hline
     $99.8$ $\%$ & $74.25_{\,2.14}$ & $47.30_{\,2.30}$ & $45.54_{\,1.79}$ & $32.98_{\,1.86}$ \\
     
    \hline
    \end{tabu}
    \caption{Results obtained by compressing a fully connected network with 100 hidden units 
    with magnitude-based pruning with $\mathbf{L2}$ regularization. 
     The standard deviation of the accuracies over independent runs (5 runs each) is given 
    as a subscript to each entry.}
    \label{tab:magl2_100}
\end{table}{}

%% file: tables/MagL1_30nhu.tex
\begin{table}[t]
    \centering
    \begin{tabu} [ 0.4\textwidth] { | X[c] | X[c] | X[c] | X[c] | X[c] |}  
     \hline
     \multicolumn{5}{ | c | }{MagL1 30 hidden units one hidden layer}\\
     \hline
     edges removed & mnist & mnist background images & mnist background random & mnist rotation\\
     \hline
     $90$ $\%$   & $95.87_{\,0.18}$ & $78.30_{\,0.59}$ & $85.97_{\,0.19}$ & $80.62_{\,0.59}$ \\ \hline
     $95$ $\%$   & $95.15_{\,0.15}$ & $77.72_{\,0.23}$ & $83.96_{\,0.12}$ & $75.99_{\,0.85}$ \\ \hline
     $97$ $\%$   & $94.38_{\,0.06}$ & $76.40_{\,0.29}$ & $80.64_{\,0.62}$ & $70.85_{\,0.87}$ \\ \hline
     $98$ $\%$   & $93.28_{\,0.11}$ & $73.54_{\,0.62}$ & $76.78_{\,0.61}$ & $65.20_{\,0.67}$ \\ \hline
     $98.5$ $\%$ & $91.71_{\,0.12}$ & $70.89_{\,0.54}$ & $72.51_{\,0.91}$ & $58.34_{\,2.05}$ \\ \hline
     $98.8$ $\%$ & $90.04_{\,0.53}$ & $68.03_{\,0.53}$ & $69.09_{\,1.14}$ & $54.38_{\,0.65}$ \\ \hline
     $98.9$ $\%$ & $89.72_{\,1.13}$ & $67.02_{\,0.47}$ & $67.92_{\,1.35}$ & $51.14_{\,1.15}$ \\ \hline
     $99$ $\%$   & $88.26_{\,0.91}$ & $65.74_{\,0.34}$ & $65.46_{\,1.42}$ & $49.20_{\,0.68}$ \\ 
 
    \hline
    \end{tabu}
    \caption{Results obtained by compressing a fully connected network with 30 hidden units 
    with magnitude-based pruning with $\mathbf{L1}$ regularization. 
     The standard deviation of the accuracies over independent runs (5 runs each) is given 
    as a subscript to each entry.}
    \label{tab:magl1_30}
\end{table}{}

%% file: tables/MagL1_100nhu.tex
\begin{table}[t]
    \centering
    \begin{tabu} [ 0.4\textwidth] { | X[c] | X[c] | X[c] | X[c] | X[c] |}  
         \hline
         \multicolumn{5}{ | c | }{MagL1 100 hidden units one hidden layer}\\
         \hline
         edges removed & mnist & mnist background images & mnist background random & mnist rotation\\
         \hline
         $0$  $\%$   & $96.77_{\,0.11}$ & $80.72_{\,0.64}$ & $88.07_{\,0.61}$ & $84.56_{\,0.29}$ \\\hline
         $50$ $\%$   & $96.77_{\,0.02}$ & $81.48_{\,0.61}$ & $88.42_{\,0.24}$ & $84.90_{\,0.85}$ \\\hline
         $90$ $\%$   & $96.43_{\,0.15}$ & $81.88_{\,0.43}$ & $88.29_{\,0.37}$ & $85.15_{\,0.95}$ \\\hline
         $95$ $\%$   & $96.32_{\,0.16}$ & $81.38_{\,0.28}$ & $88.42_{\,0.21}$ & $84.05_{\,0.76}$ \\\hline
         $97$ $\%$   & $96.09_{\,0.13}$ & $81.06_{\,0.27}$ & $87.71_{\,0.41}$ & $82.40_{\,0.45}$ \\\hline
         $98$ $\%$   & $95.70_{\,0.26}$ & $80.77_{\,0.52}$ & $86.24_{\,0.32}$ & $79.27_{\,0.75}$ \\\hline
         $98.5$ $\%$ & $95.32_{\,0.09}$ & $79.76_{\,0.40}$ & $84.34_{\,0.28}$ & $76.81_{\,0.76}$ \\\hline
         $98.8$ $\%$ & $94.92_{\,0.18}$ & $78.58_{\,0.58}$ & $82.50_{\,0.37}$ & $73.51_{\,0.64}$ \\\hline
         $98.9$ $\%$ & $94.61_{\,0.19}$ & $78.16_{\,0.30}$ & $81.84_{\,0.62}$ & $72.13_{\,0.45}$ \\\hline
         $99$ $\%$   & $94.31_{\,0.20}$ & $77.85_{\,0.24}$ & $80.83_{\,0.29}$ & $70.44_{\,1.05}$ \\\hline
         $99.3$ $\%$ & $93.46_{\,0.32}$ & $73.36_{\,0.59}$ & $76.12_{\,0.54}$ & $62.54_{\,1.14}$ \\\hline
         $99.5$ $\%$ & $91.39_{\,0.12}$ & $68.97_{\,0.76}$ & $68.39_{\,2.64}$ & $53.45_{\,1.18}$ \\\hline
         $99.7$ $\%$ & $84.37_{\,0.94}$ & $56.21_{\,1.23}$ & $51.47_{\,1.04}$ & $37.63_{\,0.92}$ \\\hline
         $99.8$ $\%$ & $65.59_{\,4.61}$ & $44.03_{\,4.41}$ & $31.08_{\,4.09}$ & $21.78_{\,5.52}$ \\\hline
    \end{tabu}
    \caption{Results obtained by compressing a fully connected network with 100 hidden units 
    with magnitude-based pruning with $\mathbf{L1}$ regularization. 
    The standard deviation of the accuracies over independent runs (5 runs each) is given 
    as a subscript to each entry.}
    \label{tab:magl1_100}
\end{table}{}

%% file: tables/MagL2_30nhu.tex
\begin{table}[t]
    \centering
    \begin{tabu} [ 0.4\textwidth] { | X[c] | X[c] | X[c] | X[c] | X[c] |} 
         \hline
         \multicolumn{5}{ | c | }{MagL2 30 hidden units one hidden layer}\\
         \hline
         edges removed & mnist & mnist background images & mnist background random & mnist rotation\\
         \hline
         $90$ $\%$   & $95.77_{\,0.08}$ & $75.52_{\,0.74}$ & $82.69_{\,0.23}$ & $78.99_{\,0.30}$ \\ \hline
         $95$ $\%$   & $94.98_{\,0.08}$ & $75.03_{\,0.36}$ & $81.31_{\,1.16}$ & $75.70_{\,0.49}$ \\ \hline
         $97$ $\%$   & $94.27_{\,0.26}$ & $74.12_{\,0.53}$ & $79.25_{\,0.49}$ & $71.21_{\,0.58}$ \\ \hline
         $98$ $\%$   & $93.20_{\,0.11}$ & $72.84_{\,0.29}$ & $76.52_{\,0.55}$ & $65.97_{\,0.74}$ \\ \hline
         $98.5$ $\%$ & $91.61_{\,0.48}$ & $70.23_{\,0.68}$ & $73.77_{\,0.16}$ & $59.56_{\,0.90}$ \\ \hline
         $98.8$ $\%$ & $90.10_{\,0.21}$ & $68.88_{\,0.37}$ & $70.73_{\,0.77}$ & $56.52_{\,0.96}$ \\ \hline
         $98.9$ $\%$ & $89.58_{\,0.08}$ & $67.34_{\,1.11}$ & $70.19_{\,0.52}$ & $53.34_{\,1.03}$ \\ \hline
         $99$ $\%$   & $87.89_{\,0.53}$ & $66.16_{\,0.47}$ & $68.06_{\,0.59}$ & $51.70_{\,0.48}$ \\ 
         \hline
    \end{tabu}
    \caption{Results obtained by compressing a fully connected network with 30 hidden units 
    with magnitude-based pruning with $\mathbf{L2}$ regularization.The standard deviation of the accuracies over independent runs (5 runs each) is given as a subscript to each entry.}
    \label{tab:magl2_30}
\end{table}{}

%% file: tables/OBD_100nhu.tex
\begin{table}[t]
    \centering
    \begin{tabu} [ 0.5\textwidth] { | X[c] | X[c] | X[c] | X[c] | X[c] |} 
         \hline
         \multicolumn{5}{ | c | }{OBD 100 hidden units one hidden layer}\\
         \hline
         edges removed & mnist & mnist background images & mnist background random & mnist rotation\\
         \hline
         $0$  $\%$   & $97.71_{\,0.05}$ & $82.29_{\,0.56}$ & $88.65_{\,0.19}$ & $88.39_{\,0.11}$ \\\hline
         $50$ $\%$   & $97.75_{\,0.09}$ & $82.17_{\,0.89}$ & $88.54_{\,0.27}$ & $88.72_{\,0.35}$ \\\hline
         $90$ $\%$   & $96.85_{\,0.22}$ & $80.79_{\,0.43}$ & $86.99_{\,1.50}$ & $86.88_{\,0.88}$ \\\hline
         $95$ $\%$   & $96.22_{\,0.20}$ & $79.85_{\,1.89}$ & $86.00_{\,1.24}$ & $84.72_{\,0.26}$ \\\hline
         $97$ $\%$   & $95.43_{\,0.24}$ & $80.80_{\,0.10}$ & $85.52_{\,1.79}$ & $81.95_{\,0.46}$ \\\hline
         $98$ $\%$   & $95.03_{\,0.13}$ & $78.95_{\,0.19}$ & $85.01_{\,0.01}$ & $77.95_{\,0.20}$ \\\hline
         $98.5$ $\%$ & $94.55_{\,0.22}$ & $78.30_{\,0.53}$ & $82.95_{\,0.34}$ & $74.09_{\,0.57}$ \\\hline
         $98.8$ $\%$ & $94.17_{\,0.15}$ & $77.04_{\,0.21}$ & $78.62_{\,1.62}$ & $70.61_{\,0.44}$ \\\hline
         $98.9$ $\%$ & $93.78_{\,0.16}$ & $76.17_{\,0.71}$ & $79.23_{\,2.45}$ & $69.19_{\,1.06}$ \\\hline
         $99$ $\%$   & $93.47_{\,0.17}$ & $75.92_{\,0.13}$ & $77.19_{\,1.88}$ & $67.49_{\,0.88}$ \\\hline
         $99.3$ $\%$ & $91.03_{\,0.78}$ & $72.32_{\,0.57}$ & $75.16_{\,1.74}$ & $59.09_{\,1.17}$ \\\hline
         $99.5$ $\%$ & $87.38_{\,1.42}$ & $68.41_{\,0.64}$ & $69.28_{\,2.67}$ & $52.44_{\,2.38}$ \\\hline
         $99.7$ $\%$ & $78.59_{\,1.57}$ & $61.18_{\,0.55}$ & $61.06_{\,1.93}$ & $41.17_{\,1.43}$ \\\hline
         $99.8$ $\%$ & $64.71_{\,5.44}$ & $53.95_{\,1.08}$ & $52.03_{\,2.32}$ & $33.77_{\,1.98}$ \\\hline
    \end{tabu}
    \caption{Results obtained by compressing a fully connected network with 100 hidden units 
    with the Optimal Brain Damage method. 
    The standard deviation of the accuracies over independent runs (5 runs each) is given 
    as a subscript to each entry.}
    \label{tab:OBD_100}
\end{table}{}

%% file: tables/OBD_30nhu.tex
\begin{table}[t]
    \centering
    \begin{tabu} [ 0.4\textwidth] { | X[c] | X[c] | X[c] | X[c] | X[c] |} 
         \hline
         \multicolumn{5}{ | c | }{OBD 30 hidden units one hidden layer} 
         \\
         \hline
         edges removed & mnist & mnist background images & mnist background random & mnist rotation\\
         \hline
         $90$ $\%$   & $95.48_{\,0.08}$ & $76.12_{\,0.81}$ & $82.28_{\,0.54}$ & $78.55_{\,0.86}$ \\ \hline
         $95$ $\%$   & $94.69_{\,0.08}$ & $74.88_{\,0.39}$ & $80.75_{\,0.59}$ & $76.16_{\,0.50}$ \\ \hline
         $97$ $\%$   & $94.08_{\,0.22}$ & $73.46_{\,0.38}$ & $79.06_{\,0.75}$ & $70.75_{\,0.42}$ \\ \hline
         $98$ $\%$   & $92.72_{\,0.38}$ & $71.69_{\,0.44}$ & $75.99_{\,0.86}$ & $64.04_{\,0.89}$ \\ \hline
         $98.5$ $\%$ & $90.53_{\,0.47}$ & $70.00_{\,0.29}$ & $73.66_{\,0.24}$ & $58.61_{\,0.06}$ \\ \hline
         $98.8$ $\%$ & $88.51_{\,0.46}$ & $68.14_{\,0.41}$ & $70.08_{\,0.15}$ & $52.93_{\,0.79}$ \\ \hline
         $98.9$ $\%$ & $86.97_{\,0.75}$ & $66.84_{\,0.36}$ & $69.51_{\,0.57}$ & $52.09_{\,0.55}$ \\ \hline
         $99$ $\%$   & $85.91_{\,0.54}$ & $65.72_{\,0.61}$ & $67.82_{\,0.45}$ & $50.71_{\,0.50}$ \\ 
        \hline
    \end{tabu}
    \caption{Results obtained by compressing a fully connected network with 30 hidden units with the
    Optimal Brain Damage method. 
    The standard deviation of the accuracies over independent runs (5 runs each) is given 
    as a subscript to each entry.}
    \label{tab:OBD_30}
\end{table}{}

%% file: figs/hist_230.tex
\begin{figure}[t]
    \includegraphics[width=1\textwidth]{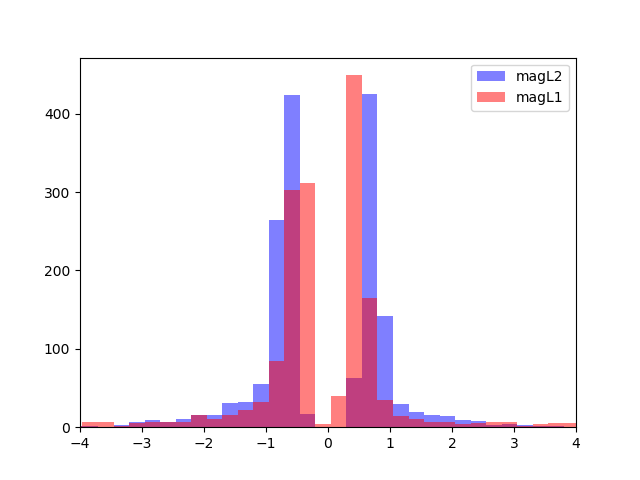}
    \caption{Histograms of the weight distribution in two networks with $50$ hidden units and compression ratio of $0.04$.
    The two networks have been pruned with magnitude-based pruning and the only difference is the regularization method.
    The histograms are obtained after 230 epochs, with accuracy above 95\% on the MNIST dataset. 
    The last pruning has been performed in epoch 210, when the network has been close to a local minimum. 
    The L1 regularization appears to have caused a stronger concentration of the weights around zero. }
    \label{fig:histogram}
\end{figure}{}

%% file: trunk/related.tex
\section{Related Work}

In this section we provide an overview on the works that concern the main 
neural-network compression techniques. 
Before discussing the individual families of methods, 
we observe that the general justification for the search of efficient 
neural-network compression methods lies in the empirical observation
that for natural tasks most network parameters appear to be redundant, 
as they can be accurately predicted from a small fraction of them
\cite{denil_predicting_2013}. 
We also note that a famous approach to reduce the number of network parameters which we don't consider in this work is that of \emph{Dark Knowledge} \cite{hinton_distilling_2015}. 
We exclude such technique based on the fact that \cite{chen_compressing_2015} shows that the method doesn't match the performance of their HashedNet; 
moreover, while \cite{chen_compressing_2015} combines HashedNet with
Dark Knowledge, as we discussed in the Experiment section, in this work
we avoid combining different methods.

\subsection{Hashing-based Techniques}

Hashing-based techniques for neural network compression have been introduced in \cite{chen_compressing_2015}, 
which proposed the \emph{HashedNet} architecture.
The idea of the technique is to randomly hash 
the network edges into \emph{buckets} and then constrain all edges between the same bucket into having the same weight. 
They can thus be naturally interpreted as weight-sharing schemes, and regarded as dimensionality reduction steps 
between network layers such as feature hashing \cite{chen_compressing_2015} and random linear sketching \cite{lin_towards_2019}. 
We remark that, in \cite{chen_compressing_2015} and in later works, HashedNet has not been directly compared to iterative
pruning techniques.
Recently, \cite{lin_towards_2019} provided rigorous theoretical results on the convexity of the 
optimization landscape of one-hidden-layer HashedNet, showing that it is similar to that
of a simple (one-hidden-layer) fully-connected network.

\cite{chen_compressing_2016} have successively applied the HashedNet technique to the frequency domain,
by first converting filter weights to the frequency domain with a discrete cosine transform before applying the HashedNet technique. 

While HashedNet and many of its variants initially group edges in different buckets before training, in unrestricted-server settings, 
\cite{hu_hashing_2018} shows that an improved hashing scheme can be computed, in order to obtain the best approximation of an
uncompressed network trained for the given task. 
In particular, \cite{hu_hashing_2018} adapts previous results for computing 
inner-product-preserving hashing functions in order
to obtain a good binary-weight approximation of the given neural network.

\subsection{Pruning-based Techniques}

An iterative pruning methods is generally identified by its measure of edge saliency, 
i.e. a way to associate a \emph{value} to individual edges so that edges with the lowest
value should be pruned first, and its \emph{pruning schedule}, that is way to decide when the pruning
should happen and how many edges should be pruned \cite{to_prune_or_not_to}. 

\subsubsection{Optimal Brain Damage}

The earliest, famous example of pruning method is Optimal Brain Damage (OBD) \cite{le_cun_optimal_1989}. 
Starting from the consideration that the pruning operation should attempt to 
increase the value of the loss function by the least possible amount, 
OBD proposes the following heuristic to determine edge saliencies: 
the loss function is first approximated via its Taylor expansion up to the 
second order, thus leading to a formula which include its gradient and its Hessian; 
first, in order to be able to ignore the linear factor involving the gradient,
OBD sets the pruning to take place when the training appears to have reached a 
local minimum;
then, since the Hessian would be unfeasible to estimate, it crucially assumes that
the extra-diagonal entries of the Hessian matrix can be assumed to be zero. 
for a formal presentation of the above procedure, 
we defer the reader to the original paper \cite{le_cun_optimal_1989} . 
Thanks to the above assumption, OBD is able to compute saliency efficiently. 
In the original paper \cite{le_cun_optimal_1989}, it is shown that OBD
leads to a much better accuracy than a simpler magnitude-based method. 

Successively, \cite{hassibi_optimal_1993} and \cite{hassibi_optimal_1994}
proposed Optimal Brain Surgeon (OBS). 
They argue that OBD's assumption of zero extra-diagonal entries for the Hessian 
fails to be satisfied in typical tasks, and they base their OBS on an exact 
expression for the saliency drop due to an edge deletion; 
however, their method reintroduce a quadratic time complexity for estimating
the Hessian, and is thus unfeasible for non-small networks.

\subsubsection{Magnitude-based Pruning}

Magnitude-based pruning methods simply define the saliency of an edge to be
its absolute value, and have recently gained popularity as a simple yet 
effective neural-network compression method \cite{to_prune_or_not_to,alford_pruned_2018}, 
also thanks to its versatility in combining with other methods \cite{han_deep_2015,han_learning_2015}.
In particular, \cite{to_prune_or_not_to} consider a magnitude-based pruning method\footnote{We
have been unable to determine whether the magnitude-based pruning considered in 
\cite{to_prune_or_not_to} make use of L1 or L2 regularization.} that
at the $i$th step of pruning reaches a sparsity level of $r(1-(1-\frac ik)^3)$, 
where $r$ is the desired compression ratio and $k$ is the total number of 
pruning steps, which is a pruning schedule which bears some similarity to those 
adopted in our pruning methods.

We remark again that, although recent works investigating magnitude-based pruning 
techniques mention OBD as a related method, they fail to test the latter 
against their method. 

%% file: trunk/compression.tex
\section{Description of Compression Methods}

In this section we describe in details the neural network compression methods 
that we experimented with. 

\subsection{Description of HashedNet}

Our experiments include the HashedNet architecture, introduced by \cite{chen_compressing_2015}. 
We performed our experiment on HashedNet by running the original code provided by the authors\footnote{The 
code can be found at \url{http://web.archive.org/web/20190902141837/https://www.cse.wustl.edu/~ychen/HashedNets/}}. 
The results of our experiments are shown in \ref{fig:old_mnist_convolution_vs_hashed}. 
As we discuss in Discussion of Convolutional Networks, carefully chosen convolutional architectures  
turn out to outperform this method across all compression ratios. 

In the HashedNet architecture with $m$ edges and compression ratio $\frac mk$,
the network is initialized by assigning, for each layer, each edge to one out of $k$ buckets, 
and by constraining the edges assigned to the same bucket to always have the same weight. 
The assignment is implemented using a hash function $f((i,j))$, 
where edge $(i,j)$ refers to the edge connecting nodes $i$ and $j$ in next and previous layers. 
In order to evaluate HashedNet, the weight of each edge $(i,j)$ is retrieved by looking up the weight
associated to its corresponding bucket $f((i,j))$.
As for training the HashedNet architecture, as it turns out by the back-propagation algorithm, 
the gradient of a bucket can be computed by summing up the gradients that all the edges assigned 
to the given bucket would have had, if they were allowed to have distinct weights.

\subsection{Description of Pruning Methods}\label{sec:pruning_description}

In (iterative) pruning methods, the desired compression level is achieved in multiple pruning steps. 
Each pruning step is performed every given number of retraining epochs from the last pruning step, 
after which edges with the least \emph{saliency} are removed.
The adopted measure of saliency of an edge depends on the pruning method under consideration.

More specifically in \emph{OBD} \cite{le_cun_optimal_1989}, 
the saliency of an edge with weight $w$ is defined to be $w^{2}\frac{\partial^2 L}{\partial w^2}$
(see the Related Work section for further details on such rule). 
In \emph{magnitude-based} pruning methods (here, \magone and \magtwo), 
the saliency of an edge $w$ is simply $|w|$.

After terminating the pruning steps by reaching the desired sparsity level, 
the network is trained for additional epochs until the accuracy reaches a local maximum.

\subsubsection{Insights on the Pruning Schedule}

Compared to pruning big chunks of edges at once, 
the purpose of reaching the desired compression by progressively pruning edges 
in multiple steps is to avoid sudden and unrecoverable accuracy drops \cite{le_cun_optimal_1989,to_prune_or_not_to}.
However, while pruning very few edges at every step would help 
in reducing the deterioration of the network accuracy, 
it would require an excessive increase in the number of epochs. 
On the one hand, bigger chunks of weights could be safely pruned 
as long as the network is above a certain level of edge density; 
on the other hand, sparse networks turns out to be more sensitive to the removal of edges, 
which suggests that pruning should be performed more gradually when the edge density becomes low. 
We thus employ iterative pruning methods which prune many edges in early phases of the training process
and less edges in later ones, and which continue training the network for a certain number of epochs
before and after each \emph{pruning phase}.

Formally, at the $i$th step of pruning our methods reach a sparsity level of $\sqrt[c]{\frac{i}{k}}S$, 
where $S$ is the final desired sparsity, $k$ is the total number of pruning steps and $c\geq 1$ is a constant. 
% Obviously, $k$ determines how smooth we converge to the desired compression and $c$ determines how much the number of removed edges reduce in every step.
%
As a consequence, in the later pruning steps, less edges are removed (the exact amount is controlled by $c$),
and we have less dramatic accuracy drops (this is controlled by $k$). 
In the Related Work section, we mention a similar method for iterative pruning 
that has been used in \cite{to_prune_or_not_to}. 
We remark that, in our experiments, we choose our parameter $c$ such 
that, after the very few initial iterations, a much higher level of sparsity is reached compared to \cite{to_prune_or_not_to}. 
The latter choice is motivated by the observation that,  
after the initialization of the weights according to a normal distribution with zero mean, 
the weight distribution has the tendency to remain normal with the same mean, with a strong concentration around zero;
moreover, after several pruning steps, the weights tend to be less concentrated around zero, as can be seen in the 
histograms in Figure \ref{fig:histogram}. 

%% file: trunk/experiments.tex
\section{Experiments}\label{section:experiment}

Experiments are carried out on the MNIST dataset and some of its variants %\cite{larochelle_empirical_2007}, 
with $50000$ training images and $12000$ test images\footnote{Note that in \cite{larochelle_empirical_2007} 
they actually employed the set of $12000$ images as training set and the set of $50000$ images as test set, against common practices on the ratio between their relative size \cite{guyonscaling}, without justifying this choice; hence, here we switched the use of the two sets.}. 
We varied the compression ratios in the range $[0.002,0.1]$,
and we tested simple architectures consisting of $100$ and $30$ hidden neurons within a single hidden layer. 
Of course for networks with $30$ neurons we did not pursue a compression of less than $0.01$, as it produced networks with very few parameters and very low accuracies that we did not find worth considering.
Each set of parameters has been tested over 5 independent runs, 
in order to reduce the variability due to the random initialization of the weights 
and to the stochastic gradient descent algorithm\footnote{The code will be made available for full reproducibility.}.

The average accuracies with their standard deviation can be seen in the tables from \ref{tab:magl2_100} to \ref{tab:OBD_30}. 
To save space, we don't include the rows corresponding to experiments with 30 hidden units 
and very low or very high compression, as the accuracy in those cases is very low and thus 
poorly informative. 

In all experiments the magnitude-based \emph{pruning} methods outperform the OBD and Convolution networks are beaten by all \emph{pruning} methods with the same number of parameters (\ref{fig:mnist_basic_100nhu}). 
Also, among the other methods, the \emph{magnitude-based} pruning with L1 regularization 
seems to outperform all others; this could be the result of the \emph{sparsification} 
effect produced by the L1 regularization \cite[Section 7.1.2]{goodfellow_deep_2016}.

\subsection{Experiments on Simple Convolutional Networks}

We compare compressing methods to a simple convolution network with three hidden layers: a 2D convolution layer followed by a max pooling layer and a final linear layer. 
We try various values for the kernel size (from 5x5 to 10x10), the number of output channels (from 4 to 12), the stride (from 1 to 3) and the pool size (from 2 to 4). 
We use a batch size of 16, a learning rate of 1e-2 and a L2 regularization with weight decay of 1e-3. We also try dropout between the max pooling layer and the linear layer (with probability from 0 to 0.2). 
The input is also normalized according to the mean and standard deviation of input values in the training set.
For each set of parameters, we make 5 runs of training with 30 epochs and compute the average test accuracy at that point as well as the number of weights in the network. 
Among all our experiments, we show only those which are not Pareto dominated with respect to these two parameters (we ignore a set of parameters if another set of parameters provides better test accuracy with less weights).

We next give a rough description of the parameters yielding the best trade-offs of test accuracy 
versus number of weights. 
For highest accuracy, a larger number of output channels is used, 
and for most of the datasets, larger convolution size also. 
For reducing the number of weights, it seems better to increase the pool size. 
And only for very scarce number of weights, it becomes interesting to increase the stride. 
Larger dropout is used in the high accuracy regime while dropout does not seem beneficial in the low accuracy regime.

\subsubsection{Remarks on the Experimental Setup}

In this section we comment about some design choices 
undertaken in carrying on our experimental evaluation, 
such as the experimentation on shallow architectures and 
the fact that we avoid testing combinations of the 
compression methods we consider. 

As for the choice of testing shallow architectures, 
their simpler structure allows for an easier investigation
of the effect of compression techniques; 
moreover, while deeper architecture have proven much 
generally more efficient in later years, there is still
an ongoing debate as for the necessity of many layers
\cite{ba_deep_2013}. 

Secondly, we notice that several related work argue 
that their method can be combined with the other ones. 
We remark that, while there is no a-priori argument 
which precludes the possibility that a combination of known techniques would 
outperform each single method applied individually, 
for the sake of understanding the efficacy of each approach
we believe that is first necessary to test the limit 
of each neural-network compression technique \emph{per-se}.
Investigating combinations of different compression methods
is surely an interesting direction for future work. 

%% file: trunk/discussion.tex
\section{Discussion}\label{sec:discussion}

In this section we 
discuss the results obtained from our experiments, 
which are represented in figures \ref{fig:old_mnist_convolution_vs_hashed} 
to \ref{fig:mnist_background_images_100nhu} (in the figures we delimit 
the area of major interest with two vertical lines, since the high-compression
regime suffers from poor accuracy and the low-compression regime 
is still uninformative as for the efficiency of the method).

\subsection{HashedNet vs Convolutional Networks}\label{ssec:conv}

A convolution layer can be seen as a hidden layer with many units and a lot of weight sharing: 
for a given output channel, the same kernel matrix is applied at various positions on the input data. 
The output for each position is thus equivalent to a unit with input degree equal to the size of the kernel matrix. 
Of course the efficiency of a convolution layer relies on known structural properties of the input 
(such as the order of the pixels in an image) whereas general compression techniques do not make any assumption on the input. 
However, it is natural to compare such general methods to a network based on a 2D convolution layer when considering datasets based on images.

\subsection{Comparison of Pruning Methods}

As discussed in the Related Work section,
OBD make use of a second order Taylor approximation of the loss function 
in order to approximate the saliency (i.e. the accuracy drop that results 
from setting weight $w$ to $0$). 
However, the aforementioned Taylor approximation requires 
the calculation of the Hessian matrix of the loss function, 
which takes time order of $O (m^2)$, where $m$ is the number of weights.
The latter would thus be a prohibitively intensive computational task. 
To overcome this issue, \cite{le_cun_optimal_1989} suggests to 
estimate the saliency of the edges by considering
only the diagonal entries of the Hessian matrix and ignoring the extra-diagonal ones, 
as this should be a good approximation when the loss function is close to a local minimum.
Magnitude-based pruning instead consists in the simple removal the edges with the least absolute value; 
the limited effect of this operation on the loss function are grounded on the Lipschitz 
property of the latter \cite{liu_sparse_2015}.
In our experiments, magnitude-based pruning consistently shows a slightly better performance than OBD 
across all data sets. 
Hence, our experiments provide concrete evidence 
that the assumptions which justify the heuristic employed by OBD fail to be satisfied, 
at least during critical parts of the training, 
corroborating the claims of other works discussed in the Related Work section. 

We also test magnitude-based pruning combined with L1 regularization. 
This approach shows to be slightly more effective than using L2 regularization 
as it can be seen in the plots \ref{fig:mnist_basic_100nhu} and \ref{fig:mnist_background_images_100nhu}. 
We argue that such improvement is due to the fact that L1 regularization tends to sparsify 
weights, forcing a larger number of weights to be close to zero \cite{goodfellow_deep_2016}:
the histograms in Fig. \ref{fig:histogram} show the weight distribution of two networks 
that have been pruned with L1 and L2 regularization, w$20$ epochs after a pruning step. 
In the figure, we can appreciate that L1 regularization 
force the weight distribution to concentrate more closely around zero.